\pdfoutput=1
\documentclass[11pt]{article}

\usepackage[preprint]{acl}

\usepackage{times}
\usepackage{latexsym}

\usepackage[T1]{fontenc}
\usepackage[utf8]{inputenc}

\usepackage{microtype}

\usepackage{inconsolata}

\usepackage{graphicx}

\title{Game Plot Design with an LLM-powered Assistant:\\An Empirical Study with Game Designers}

\author{\textbf{Seyed Hossein Alavi}$^{\thanks{This work was partially completed during the author’s internship at Microsoft.} 1, 4}$~~ \textbf{Weijia Xu}$^2$~~ \textbf{Nebojsa Jojic}$^2$~~ \textbf{Daniel Kennett}$^3$~~ \textbf{Raymond T. Ng}$^1$ \\
    \textbf{Sudha Rao}$^2$~~ \textbf{Haiyan Zhang}$^3$~~ \textbf{Bill Dolan}$^2$~~ \textbf{Vered Shwartz}$^{1, 4}$ \\
    $^1$University of British Columbia~~
    $^2$Microsoft Research~~
    $^3$Microsoft Gaming~~
    $^4$ Vector Institute for AI
    \\
    \texttt{salavis@cs.ubc.ca}
    }

\usepackage{amsmath}
\usepackage{booktabs}
\usepackage{graphics}
\usepackage{changepage}
\usepackage{graphicx, multirow}
\usepackage{amsmath}
\usepackage[linesnumbered,ruled]{algorithm2e}
\usepackage{makecell}
\usepackage{array}
\usepackage{arydshln} 
\usepackage[toc,page]{appendix}
\usepackage[export]{adjustbox}
\usepackage{tabularx}
\usepackage{tikz}
\usepackage{enumitem}

\usepackage{pgf-pie} 
\usepackage{enumerate}

\usepackage{enumitem}
\definecolor{deeppink}{rgb}{1.0, 0.08, 0.58}
\definecolor{cadmiumgreen}{rgb}{0.0, 0.42, 0.24}
\newcommand{\ourtool}{GamePlot}

\begin{document}
\maketitle

\begin{abstract}

We introduce \ourtool{}, an LLM-powered assistant that supports game designers in crafting immersive narratives for turn-based games, and allows them to test these games through a collaborative game play and refine the plot throughout the process. Our user study with 14 game designers shows high levels of both satisfaction with the generated game plots and sense of ownership over the narratives, but also reconfirms that LLM are limited in their ability to generate complex and truly innovative content. We also show that diverse user populations have different expectations from AI assistants, and encourage researchers to study how tailoring assistants to diverse user groups could potentially lead to increased job satisfaction and greater creativity and innovation over time.
\end{abstract}
\section{Introduction}
\label{sec:introduction}

The  landscape of interactive entertainment and the escalating player expectations has led to increased demand for innovative tools to support the work of game designers. In particular, the process of crafting compelling narratives within games can be labor intensive, and maintaining coherence and engagement throughout the game can be challenging \cite[][]{10.5555/940352}. 

Large language models \cite[LLMs;][]{radford2019language} hold promise as a support tool to augment and enhance the manual process of game design. Previous work used LLMs to generate dialogues between players and non-player characters  \cite[NPCs;][]{volum-etal-2022-craft}, to facilitate player-driven creation of new elements in the game world \citep{10333153}, and to help players uncover new narrative paths in a text-based games \cite{peng2024player-driven}, among others \cite{10.1145/3640794.3665582}. 

\begin{figure*}[t]
    \centering
\includegraphics[width=.9\textwidth,trim={0 0 0 2cm},clip]{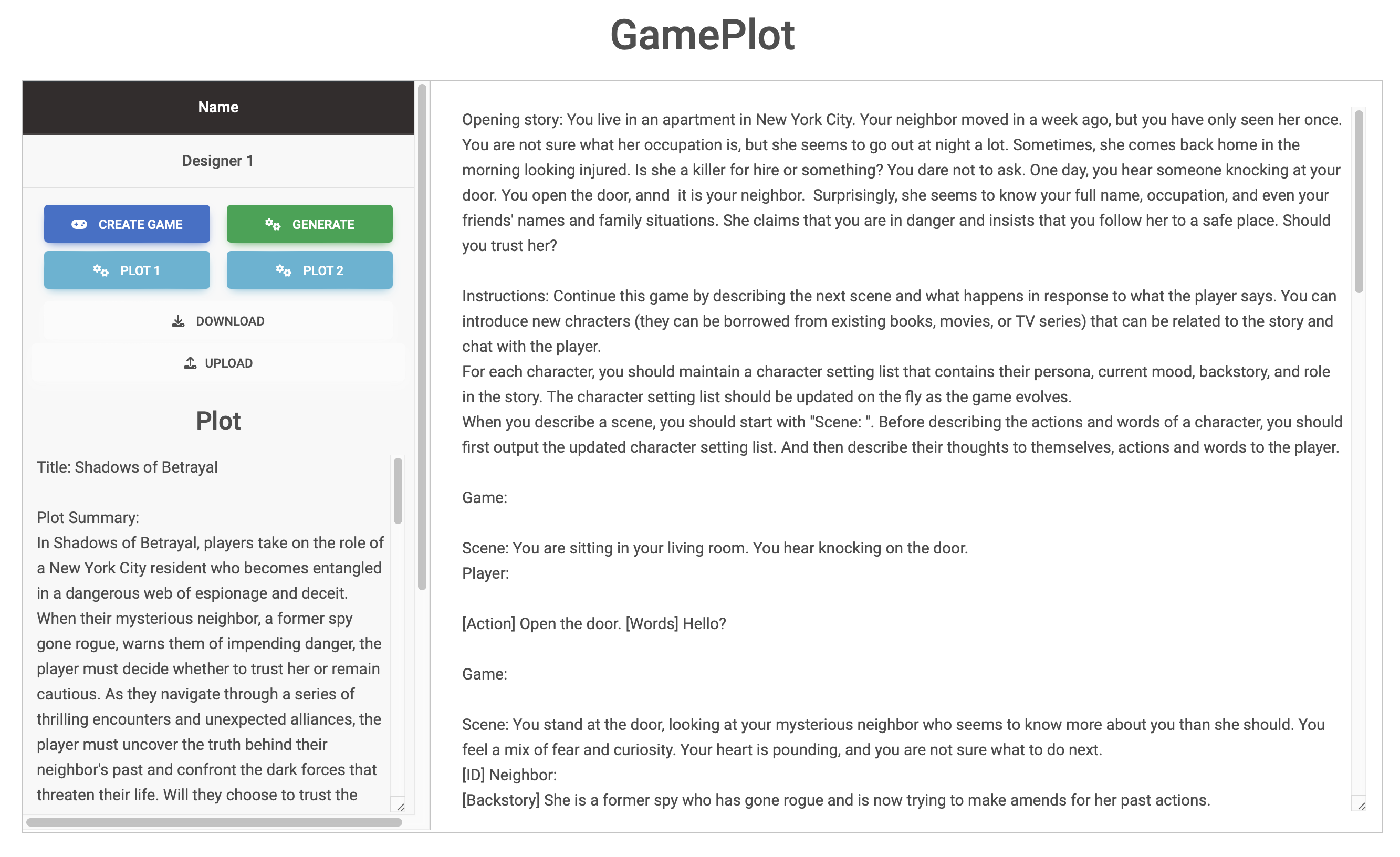}
    \caption{An illustration of the \ourtool{} design room. The right pane shows the opening story, LLM instructions, and game and player turns as designers refine the narrative. The left pane includes buttons for generating story content, creating plots, saving and loading progress, and editable sections for plot and feedback features (located below the plot area).} 
    \label{fig_designroom}
\end{figure*}

In this paper, we study the efficacy of LLMs in game narrative design. We introduce \ourtool{} (Fig.~\ref{fig_designroom}), an LLM-based web application designed to support game designers of any skill level in the process of game narrative development. Prior work showed that interactions with LLMs can lead to emergent storytelling opportunities, enhancing player engagement and creativity \cite{peng2024player-driven}. Drawing on these findings, we designed \ourtool{} to support game designers in \textbf{generating} and \textbf{refining} narratives. Beyond the use of LLMs in the initial game development, \ourtool{} also offers a game room that enables real-time narrative refinement based on player interactions and feedback, while maintaining narrative coherence and quality. Additionally, it features a Wizard of Oz (WOZ) functionality, allowing designers to discreetly assume control of NPCs and interact with players directly as part of the development phase. Overall, \ourtool{} was developed to empower game designers to create intricate storylines, develop dynamic and evolving NPCs, and shape game scenes and settings with ease and flexibility.

To assess the efficacy of LLMs in assisting game designers using \ourtool{}, we conducted a user study, inviting 14 game developers and narrative designers to engage with the tool and share their feedback. Our findings indicate a positive reception among participants. Participants were satisfied with the generated game plots, considered the AI assistant as beneficial in enhancing the game storytelling, and also reported a sense of control and ownership over the narratives. Furthermore, they found the game room and the iterative refinement and feedback gathering step to be very valuable, particularly enjoying the ability to dynamically adjust the storyline during testing sessions. 

By analyzing individual responses, we observed that game developers often prefer to offload the narrative writing process to the AI, whereas narrative writers would rather maintain creative control over the narrative but can benefit from the AI's assistance in exploring narrative paths. Yet, despite the utility of LLMs in this task, they are still limited in their ability to generate nuanced and complex narratives that engage experienced writers. Finally, we encourage researchers to identify groups of users and tailor future assistants to their specific needs instead of developing a ``one-size-fits-all'' assistant. This can potentially lead not only to enhanced productivity but also to job satisfaction and greater creativity and innovation over time.%\footnote{We will make our code and data available upon publication.}

\section{Background}
\label{sec:background}

\paragraph{LLMs in Gaming.} Language models have commonly been used to generate dialogues with NPCs \cite[e.g.,][]{10.1145/3472538.3472595,gao-emami-2023-turing, alavi2024mcpdialminecraftpersonadrivendialogue}. \newcite{10.1609/aiide.v19i1.27504} generated branching conversation paths with NPCs based on player choices, while \citet{akoury2023towards} used LLMs to erate contextually-grounded NPCs dialogues in video games. In a similar line of work, LLMs have been used to power agents (NPCs and players) in games \cite{Hausknecht2019InteractiveFG,wang2023voyageropenendedembodiedagent}.

In terms of narrative generation, LLMs were used for generating quests \cite{9980408} as well as interactive stories \cite{10.1145/3402942.3409599}. \citet{10.1145/3544548.3581441} used knowledge graphs and LLMs to generate personalized quests based on player-NPC dialogues and actions. Similarly, \citet{10343021} used LLMs to streamline the creation of serious games, automating the generation of adaptive learning content within narrative structures. In comparison to prior work, \ourtool{} goes beyond content generation with LLMs and facilitates a collaborative AI-designer process designers to iterate and refine the generated content. 

GENEVA \citep{leandro2024geneva} is another collaborative tool similar to \ourtool{}. It uses LLMs to generate branching narratives based on a high-level description from the designer, and visualize complex story paths through interactive graphs. In comparison, \ourtool{} enables designers to design game narratives through gameplay collaboratively with LLMs and refine them at any stage of the design process.

\paragraph{LLMs for Creative Writing.} Various LLM-powered tools have been developed to assist creative writers \cite[e.g.,][]{10.1145/3490099.3511105,10.1145/3544548.3581225,10.1145/3635636.3656201}. Wordcraft \citep{10.1145/3490099.3511105} allows writers to have open-ended conversations with an LLM about their stories and enables custom requests, such as rewriting sections and generating new story elements. Similarly, Dramatron \cite{10.1145/3544548.3581225} provides a co-writing environment for screenplays and theater scripts using LLMs. Given a short description of a movie, the tool generates a title, characters, plot outline, locations, and dialogue for each scene, allowing writers to continue the generation process, edit responses, or regenerate outputs as needed.

Our study is inspired by two studies that surveyed professional writers on their experience using creative writing AI assistants \cite{ippolito2022creativewritingaipoweredwriting,10.1145/3635636.3656201}. Both studies revealed inherent limitations with LLM-based creative writing, including the generation of repetitive, predictable, and clich\'{e}d responses, challenges in maintaining the writer's unique style and voice, and superficial language understanding abilities. Our user study (Sec~\ref{sec:evaluation}) reveals similar limitations when using LLMs for game narrative design.

\section{\ourtool}
\label{sec:GamePlot}

\ourtool{}\footnote{The code is available at: \url{https://github.com/salavi/GamePlot-LLM-Assistant}} is an AI assistant tailored for the early stages of game development, with a specific focus on story creation. 
\ourtool{} comprises two rooms: the design room (Sec~\ref{sec:gameplot:design}) and the game room (Sec~\ref{sec:gameplot:game}). Upon arriving at the landing page, designers can choose to enter the design room, where they develop the game story. 
Upon completion, the tool generates a \emph{game plot}, the summary of key events and characters, which will be used as the starting point for future game sessions. Both players and designers can enter the game room where they can play the game, provide feedback (players) and refine the plot based on player feedback (designers). We detail both rooms below, and conclude with the implementation details (Sec~\ref{sec:impl}). 

\begin{table}[t]
    \centering
    \scriptsize
    \setlength{\tabcolsep}{3.2pt}
    \tt
    \begin{tabular}{|p{0.48\textwidth}|}
    \hline
         % \textcolor{cadmiumgreen}{\#Few-shot examples}\\
        \textcolor{red}{Game:}\\
        \textcolor{blue}{Scene:} You are sitting in your living room. You hear knocking on the door.\\
         
         \textcolor{cadmiumgreen}{Player:} \\
         \textcolor{blue}{[Action]} Open the door.\\
         \textcolor{blue}{[Words]} Hello?\\
         \textcolor{red}{Game:}\\
         \textcolor{blue}{Scene:} You stand at the door, looking at your mysterious neighbor who seems to know more about you than she should. You feel a mix of fear and curiosity. Your heart is pounding, and you are not sure what to do next.\\
         \textcolor{blue}{[ID]} Neighbor:\\
         \textcolor{blue}{[Backstory]} She is a former spy who has gone rogue and is now trying to make amends for her past actions.\\
         \textcolor{blue}{[Persona]} Confident, but also a little anxious.\\
         \textcolor{blue}{[Mood]} Urgent and a little bit scared.\\
         \textcolor{blue}{[Thought]} I need to get them out of here before it's too late.\\
         \textcolor{blue}{[Action]} Takes a step forward and looks directly into your eyes.\\
         \textcolor{blue}{[Words]} ``Please, you don't have much time. I know this might sound crazy, but you are in danger. Can you trust me?''\\
         \textcolor{cadmiumgreen}{Player:}\\
    \hline
    \end{tabular}
    \caption{Initial turns of a template game story provided to the game designers in Sec~\ref{sec:evaluation}.}
    \label{tab:story_turns}
\end{table}

\subsection{Design Room}
\label{sec:gameplot:design}

The design room (Fig.~\ref{fig_designroom}) offers the following features to aid game designers.

\paragraph{Game Story Development.} The design room provides a text window for designers to develop their storylines by playing the game. The designer's initial inputs may include: (a) an opening story, which will be used by the designer to create the game and will initialize the game session for players (see Appendix~\ref{sec:app:opening}); (b) instructions to the LLM, such as ``Continue this game [...] You can introduce new characters [...]'' (see full example in Appendix~\ref{appendix.instructions}); and (c) 1-2 game and player turns, which serve as in-context examples for the LLM.

After the initial inputs, the designer proceeded to design the game through game play. They can choose whether to write the current turn or use the LLM to generate it. If the designer wrote a game turn, the LLM responds with a player turn, and vice versa. Table~\ref{tab:story_turns} demonstrates the initial turns in a game design. As can be seen, game turns can involve introducing NPCs along with their backstory, mood, persona, and more (see appendices~\ref{appendix.tags} for the tag inventory). The window is fully editable, allowing designers to modify both previous and current turns.

\begin{figure*}[t]
    \centering
    \setlength{\tabcolsep}{3.2pt}
    \begin{tabular}{|c|c|c|}
    \hline
         \includegraphics[width=.5\textwidth]{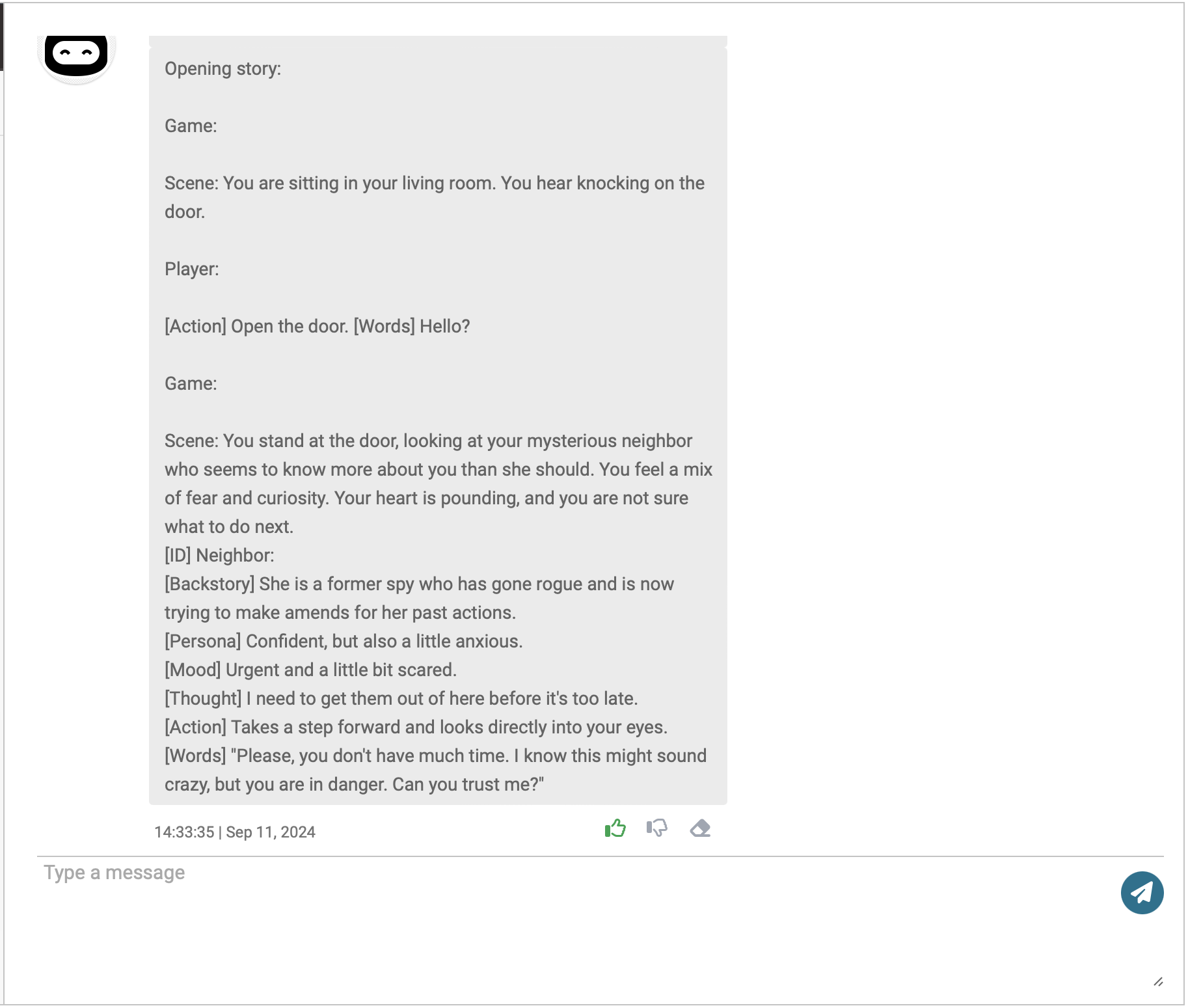} & \includegraphics[width=.21\textwidth]{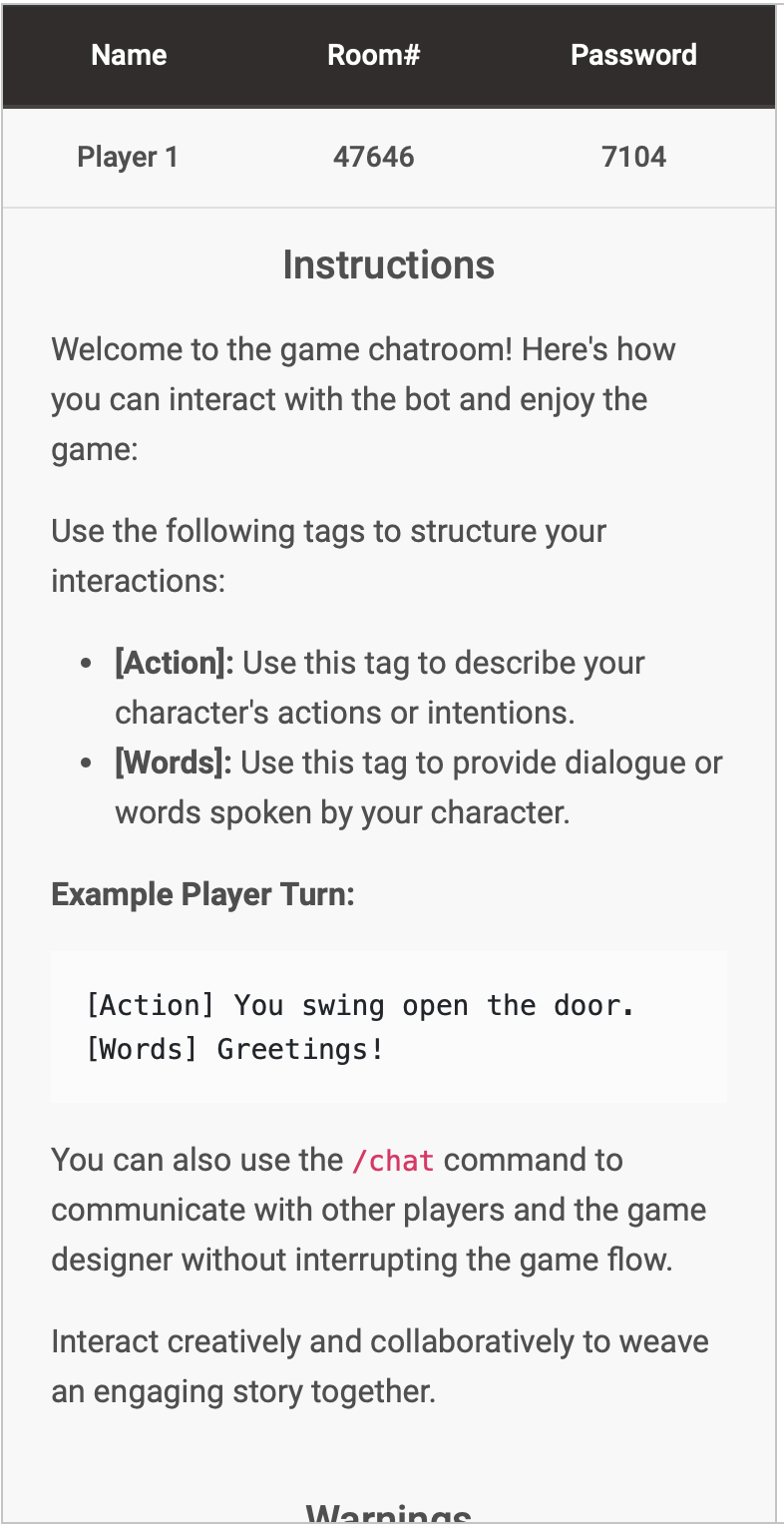} &  \includegraphics[width=.21\textwidth]{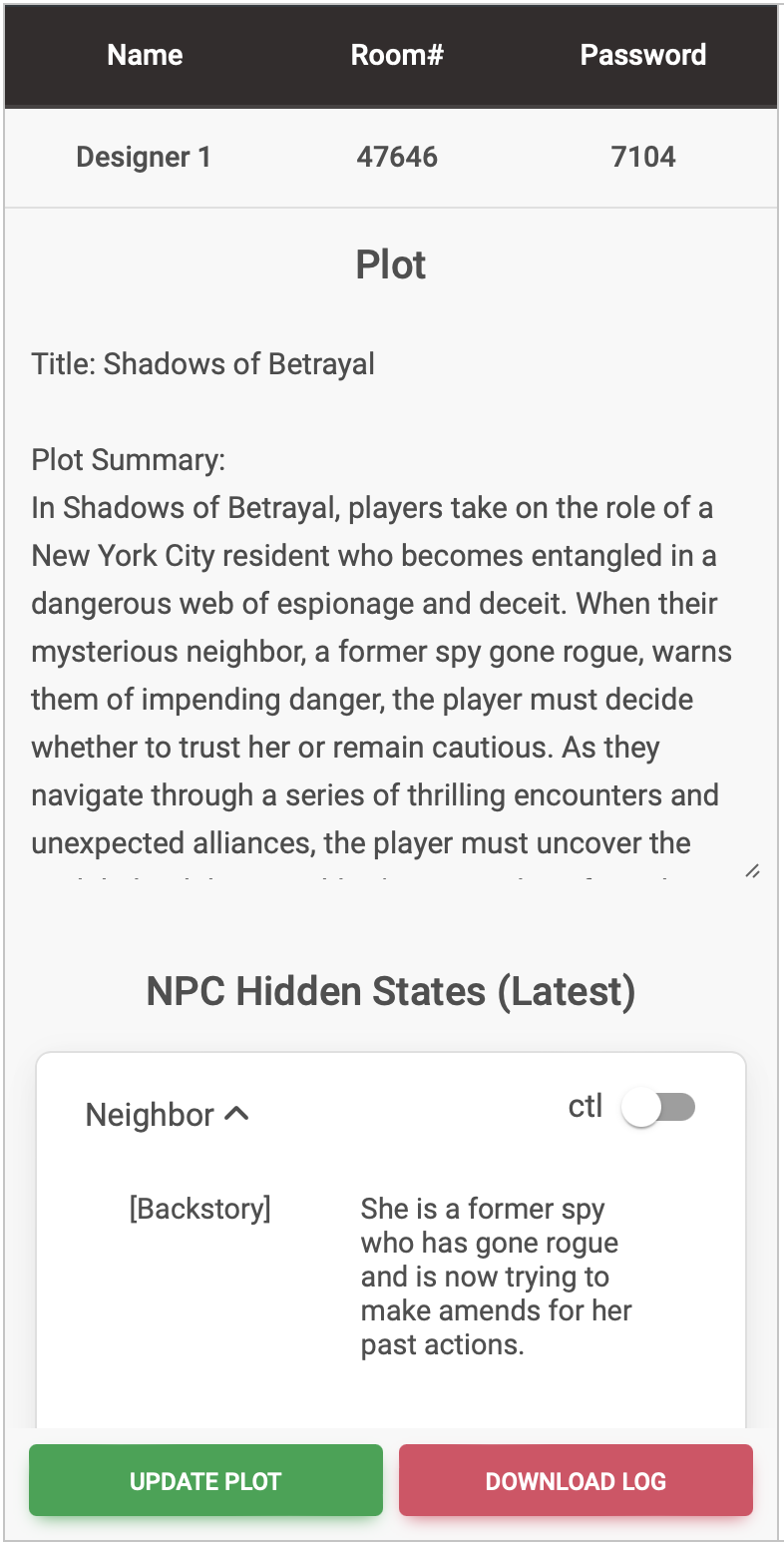} \\
         \hline
    \end{tabular}
    \caption{An illustration of the \ourtool{} game room. \textbf{Left}: game window from the designer's perspective, where the game is played. \textbf{Middle}: left pane of the player view, including room and user information, and gameplay instructions. \textbf{Right}: left pane of the designer view, including an editable text area prefilled with the plot and the list of NPCs that appeared so far in the game window. For each NPC, designers (but not players) see a list of hidden states (e.g., backstory, persona, mood, and thought). By toggling the Ctl button, designers can take control of that NPC and use the Wizard of Oz feature.}
    \label{fig_gameroom}
\end{figure*}
\begin{table}[t]
    \centering
    \scriptsize
    \tt
    \setlength{\tabcolsep}{3.2pt}
    \begin{tabular}{|p{0.48\textwidth}|}
    \hline
         % \textcolor{cadmiumgreen}{\#Few-shot examples}\\
        \textcolor{cadmiumgreen}{title:} Shadows of Betrayal\\
        \textcolor{cadmiumgreen}{Plot summary:}\\
        In Shadows of Betrayal, players take on the role of a New York City resident who becomes entangled in a dangerous web of espionage and deceit. When their mysterious neighbor, a former spy gone rogue, warns them of impending danger, the player must decide whether to trust her or remain cautious. As they navigate through a series of thrilling encounters and unexpected alliances, the player must uncover the truth behind their neighbor's past and confront the dark forces that threaten their life. Will they choose to trust the neighbor and embark on a dangerous journey, or will they rely on their own instincts to survive?\\
         
         \textcolor{cadmiumgreen}{Key Events:}\\
        1. The player opens the door to their neighbor, who claims to know personal details about them and insists they are in danger.\\
        2. The neighbor reveals her backstory as a former spy seeking redemption and urges the player to trust her.\\
        3. ...\\
        .\\
        .\\
        .\\
        \textcolor{cadmiumgreen}{NPCs:}\\
        \textcolor{blue}{[ID]} Neighbor\\
        \textcolor{blue}{[Backstory]} She has been living a life of secrecy and mistrust, making it difficult for her to gain the trust of others.\\
        \textcolor{blue}{[Persona]} Determined, but also wounded.\\
    \hline
    \end{tabular}
    \caption{An example plot summarizing one of the games designed in our study.}
    \label{tab:sample_plot}
\end{table}

\paragraph{Plot.} Once designers have fleshed out the storyline by playing through the game in the main window, they can click a button to summarize the key elements of the game into a structured \emph{plot} (see Table~\ref{tab:sample_plot} for an example). Designers can edit the generated plot, and once they finalize it, they can use it to initialize a game session in the game room.

\paragraph{Feedback Elicitation.} Designers can specify what they would like to receive feedback about from the players. This allows them to query players for feedback regarding specific aspects of concern such as ``the response aligns / doesn't align with the NPC character''.

\subsection{Game Room}
\label{sec:gameplot:game}

Once a game room is created, designers can share the room with players for collaborative gameplay, where the LLM generates the game turns, and players engage with the game through player turns. Figure~\ref{fig_gameroom} presents the game room interface. On the left, the game window is shown, where the game is played (by designers or players). The middle parts and the right part of the figure demonstrate the different actions that players and designers can perform in the room, as detailed below.

\paragraph{Game Window.} The game window displays the game and player turns. It is initialized with the plot (Sec~\ref{sec:gameplot:design}), which guides the LLMs in playing the game turns. Players can play their turns and provide feedback on the game turns. Designers can participate in the game themselves or monitor the narrative as the players see it. 

\paragraph{Designer Control.} In the game room, designers have additional controls to monitor and intervene in the game flow if necessary (Fig.~\ref{fig_gameroom}, right). 
First, as the story unfolds and players interact with the game, designers may wish to modify the plot. Designers can make live changes to key events that haven't been played yet through the game interface. 

Second, on the same pane, the designer sees a list of NPCs that have appeared in the narrative and their corresponding hidden tags (e.g., \texttt{[Mood]}, \texttt{[Thought]}, etc.). The same information is not visible to players. 

Finally, designers can assume control of an NPC. Technically, this is implemented by identifying game turns that refer to the controlled NPC and allowing designers to edit and approve them. The players are unaware that the game turns are controlled by the designer. This Wizard of Oz-style experimentation allows designers  
to adjust NPC responses, guide players out of narrative dead ends, or modify the game flow without breaking the player's immersion. 

\subsection{Implementation Details}
\label{sec:impl}

We used \texttt{GPT-3.5-Turbo-16k} as the backbone LLM for \ourtool{} due to its cost-efficiency, low response time, and good performance.   
We used the model's default hyperparameters, setting the maximum token limit to 2,000 for summarization and plot generation, and 1,000 for generating the next turn (see prompt details in Appendix \ref{app:prompts}).
To address the LLM token limit which may be exceeded in long games, once we hit the maximum token limit (40,000 characters), we keep the last k = 10 turns and use the LLM to summarize earlier turns. Lastly, we used the stop words [\texttt{``Player:'', ``Game:''}] to indicate to the LLM which turn it needs to generate. 
\section{User Study}
\label{sec:evaluation}

\begin{figure}[t]
    \centering
    \scriptsize
    \begin{tikzpicture}
    \pie[radius=1.5,rotate = 45, text = inside, color = {orange!70,green!75!black,blue!30}]
    {64.3/> 1 year,
        14.3/< 1 year,
        21.4/Hobbyist}
    \end{tikzpicture}
    \caption{Distribution of participants by experience level. The majority of participants had more than 1 year of experience, while those with less than 1 year of game design experience had significant experience in narrative design.}
    \label{fig_participants_dist}
\end{figure}
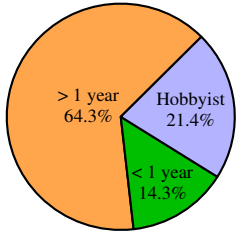

To evaluate the effectiveness of \ourtool{}, we recruited participants with backgrounds in both narrative design and game development. Participants were first given a brief tutorial on \ourtool{} and its key features. Following the tutorial, they were provided with a template story, including the premise of the game, instructions, and the first few turns (see Appendix~\ref{app:game_story_template}). Participants were asked to use the template story to design a game in the design room, for which they were given 20-25 minutes. Once they were done, the resulting game plot was used to initialize a game room, to which they logged into as designers. Along with the designer, one of the authors of this paper joined the game room as a player to collaboratively test the game with them. Participants were given 15-20 minutes to test the game. 

We detail below the recruitment process (Sec~\ref{sec:participants}). At the end of the testing sessions, participants had to fill out a post-study questionnaire about their experience. Questions were design to understand the strengths and weaknesses of the \ourtool{} (Sec~\ref{sec:feedback_gameplot}), as well as to more broadly query them about how open they were to the idea of designing game stories collaboratively with AI (Sec~\ref{sec:feedback_ai}). 

\subsection{Participants}
\label{sec:participants}

We recruited 14 participants through Upwork and social media advertisements. The majority of the participants (64.3\%) had at least one year of professional experience in the gaming industry (Fig.~\ref{fig_participants_dist}). Of the 14 participants, 8 indicated that they had expertise in narrative writing. This diverse group ensured a comprehensive evaluation of \ourtool{} from the perspectives of both narrative writers and game designers. Each session lasted between 1 to 2 hours, and participants were compensated at a rate of \$25 per hour. 

\subsection{Feedback on \ourtool{} Features}
\label{sec:feedback_gameplot}

\input{Figures/fig_features}

\paragraph{Valuable Features.} We asked the participants about the features they found most valuable in the game room. In particular, we asked them whether they liked the following features: the ability to change the plot during the test session, the NPC summary, the collaborative game play, and the feedback buttons. We also asked them which 1-3 features in \ourtool{} they found most valuable and would like to see retained, allowing them to provide free-text answers. Table~\ref{fig_features} presents the results. The top part shows the number of participants that favored each of the specific features we asked about, while the bottom part shows the number of participants that mentioned each other feature in their free-text responses. For each feature, we also present an example feedback from the participants. 

We observe that the most liked feature was the ability to change the game plot during the test session, which was favored by 12 out of 14 participants. Half of the participants also favored the NPC hidden states dropdown (NPC summary) and the multiplayer setup. When specifically asked about how useful the game room was overall, participants rated its usefulness for testing with players highly, with 13 out of 14 giving ratings of 4 or 5, resulting in an average score of 4.21.

In the free-text responses, several features of \ourtool{} stood out. Participants appreciated the ability to generate content (for example, generating multiple plots for the same story), as well as the level of control they retained and the ability to modify the generated content. Several participants mentioned that they liked the ability to go back and edit the game and have the tool propagate the changes to later turns. A key theme that emerged was \ourtool{}'s adaptability and control in story generation in the design room. Designers consistently emphasized how the ability to modify elements such as NPC moods, environments, and plot details allowed them to directly influence AI-generated responses. This dynamic adaptability proved to be a strong draw for users who sought control over the narrative direction.

Another feature that was mentioned by several of the participants was the ability to assume control over NPCs in the game room. Designers spoke highly of the ``Wizard of Oz'' setup, which enables game designers to control NPC interactions, adjust the plot, and exercise narrative control without disrupting the players' experience and their immersion in the game. 

Finally, ease of use and collaborative features also surfaced as important strengths. Participants appreciated the simplicity of the user interface, describing it as intuitive and easy to navigate. The ability to involve multiple users in testing and design further enhanced the tool's appeal;  Participants liked the idea of designing a plot and then testing it and iteratively improving it along with their team. 

\begin{table*}[t]
    \centering
    \scriptsize

\begin{tabular}{P{4.5cm}cP{8cm}}
\toprule

    \textbf{Category} & \textbf{Mentioned by} & \textbf{Examples of Feedback} \\ \midrule

    UI Improvement & \begin{tikzpicture}[baseline=-10]
    % Define the number of rows and columns
    \def\rows{2}
    \def\cols{7}

    % Define the spacing between the dots
    \def\xspacing{0.3}  % Horizontal spacing
    \def\yspacing{0.3}  % Vertical spacing

    % Counter for the dots
    \def\counter{0}

    % Loop to create the grid of dots
    \foreach \x in {1,...,\cols} {
        \foreach \y in {1,...,\rows} {
            % Increment the counter
            \pgfmathtruncatemacro{\counter}{(\y - 1) * \cols + \x}

            % Choose color based on counter value
            \ifnum\counter<10
                \fill[orange!70] (\x*\xspacing, -\y*\yspacing) circle (3pt); % First 12 dots red
            \else
                \fill[gray!70] (\x*\xspacing, -\y*\yspacing) circle (3pt); % Last 2 dots gray
            \fi
        }
    }
\end{tikzpicture} & ``Instead of having /chat inside the game, maybe there should be a chat box off to the side that players can write in so it doesn't distract the gameplay.'' \\ \midrule
    
    Enhancement & \begin{tikzpicture}[baseline=-10]
    % Define the number of rows and columns
    \def\rows{2}
    \def\cols{7}

    % Define the spacing between the dots
    \def\xspacing{0.3}  % Horizontal spacing
    \def\yspacing{0.3}  % Vertical spacing

    % Counter for the dots
    \def\counter{0}

    % Loop to create the grid of dots
    \foreach \x in {1,...,\cols} {
        \foreach \y in {1,...,\rows} {
            % Increment the counter
            \pgfmathtruncatemacro{\counter}{(\y - 1) * \cols + \x}

            % Choose color based on counter value
            \ifnum\counter<4
                \fill[orange!70] (\x*\xspacing, -\y*\yspacing) circle (3pt); % First 12 dots red
            \else
                \fill[gray!70] (\x*\xspacing, -\y*\yspacing) circle (3pt); % Last 2 dots gray
            \fi
        }
    }
\end{tikzpicture} & ``[...] present a few options of how the plot might go down when I'm in design phase.'' /  ``[...] Generating images according to the scenes.'' \\ \midrule

Ease of Use & \begin{tikzpicture}[baseline=-10]
    % Define the number of rows and columns
    \def\rows{2}
    \def\cols{7}

    % Define the spacing between the dots
    \def\xspacing{0.3}  % Horizontal spacing
    \def\yspacing{0.3}  % Vertical spacing

    % Counter for the dots
    \def\counter{0}

    % Loop to create the grid of dots
    \foreach \x in {1,...,\cols} {
        \foreach \y in {1,...,\rows} {
            % Increment the counter
            \pgfmathtruncatemacro{\counter}{(\y - 1) * \cols + \x}

            % Choose color based on counter value
            \ifnum\counter<4
                \fill[orange!70] (\x*\xspacing, -\y*\yspacing) circle (3pt); % First 12 dots red
            \else
                \fill[gray!70] (\x*\xspacing, -\y*\yspacing) circle (3pt); % Last 2 dots gray
            \fi
        }
    }
\end{tikzpicture} & ``Clearer instructions for people using this tool on all of its functionalities and features so that game developers can more easily understand and use this tool.'' \\ \midrule

Performance & \begin{tikzpicture}[baseline=-10]
    % Define the number of rows and columns
    \def\rows{2}
    \def\cols{7}

    % Define the spacing between the dots
    \def\xspacing{0.3}  % Horizontal spacing
    \def\yspacing{0.3}  % Vertical spacing

    % Counter for the dots
    \def\counter{0}

    % Loop to create the grid of dots
    \foreach \x in {1,...,\cols} {
        \foreach \y in {1,...,\rows} {
            % Increment the counter
            \pgfmathtruncatemacro{\counter}{(\y - 1) * \cols + \x}

            % Choose color based on counter value
            \ifnum\counter<2
                \fill[orange!70] (\x*\xspacing, -\y*\yspacing) circle (3pt); % First 12 dots red
            \else
                \fill[gray!70] (\x*\xspacing, -\y*\yspacing) circle (3pt); % Last 2 dots gray
            \fi
        }
    }
\end{tikzpicture} & ``Server time, it could be quicker.'' \\

\bottomrule
\end{tabular}

    \caption{Areas of improvement in \ourtool{} that were mentioned by participants in the study as free-text feedback (manually classified by the authors).}
    \label{tbl_areas_for_improvement}
\end{table*}
\begin{table*}[t]
    \centering
    \scriptsize

\begin{tabular}{P{4.5cm}cP{8cm}}
\toprule

    \textbf{Category} & \textbf{Mentioned by} & \textbf{Examples of Feedback} \\ \midrule

    Lack of Creativity & \begin{tikzpicture}[baseline=-10]

    \def\rows{2}
    \def\cols{7}

    \def\xspacing{0.3}  % Horizontal spacing
    \def\yspacing{0.3}  % Vertical spacing

    % Counter for the dots
    \def\counter{0}

    % Loop to create the grid of dots
    \foreach \x in {1,...,\cols} {
        \foreach \y in {1,...,\rows} {
            % Increment the counter
            \pgfmathtruncatemacro{\counter}{(\y - 1) * \cols + \x}

            % Choose color based on counter value
            \ifnum\counter<4
                \fill[orange!70] (\x*\xspacing, -\y*\yspacing) circle (3pt); % First 12 dots red
            \else
                \fill[gray!70] (\x*\xspacing, -\y*\yspacing) circle (3pt); % Last 2 dots gray
            \fi
        }
    }
\end{tikzpicture} & ``It is a bit linear, so it seems unlikely that the AI will come up with something really unique and interesting without a lot of input from the dev.'' \\
& & ``Can be a little straight forward in its narrative'' \\ 
& & ``Obvious answers. No surprise or engagement effect.'' \\ \midrule

Interaction with the AI &  \begin{tikzpicture}[baseline=-10]
    % Define the number of rows and columns
    \def\rows{2}
    \def\cols{7}

    % Define the spacing between the dots
    \def\xspacing{0.3}  % Horizontal spacing
    \def\yspacing{0.3}  % Vertical spacing

    % Counter for the dots
    \def\counter{0}

    % Loop to create the grid of dots
    \foreach \x in {1,...,\cols} {
        \foreach \y in {1,...,\rows} {
            % Increment the counter
            \pgfmathtruncatemacro{\counter}{(\y - 1) * \cols + \x}

            % Choose color based on counter value
            \ifnum\counter<4
                \fill[orange!70] (\x*\xspacing, -\y*\yspacing) circle (3pt); % First 12 dots red
            \else
                \fill[gray!70] (\x*\xspacing, -\y*\yspacing) circle (3pt); % Last 2 dots gray
            \fi
        }
    }
\end{tikzpicture} & ``Flow. As I writer I want to remain in a `flow state' where I can just write what comes to mind. Having to enter tags slows me down.'' \\
& & ``It should generate whole stories like we can generate using other free AI tools available online like Chatgpt, Gemini or Meta AI.'' \\ 
& & ``Learning curve, OF THE USER not the AI. Users might find it frustrating while getting started. Especially when missing simple commands.'' \\ \midrule

Quality of Generations - Other & \begin{tikzpicture}[baseline=-10]
    % Define the number of rows and columns
    \def\rows{2}
    \def\cols{7}

    % Define the spacing between the dots
    \def\xspacing{0.3}  % Horizontal spacing
    \def\yspacing{0.3}  % Vertical spacing

    % Counter for the dots
    \def\counter{0}

    % Loop to create the grid of dots
    \foreach \x in {1,...,\cols} {
        \foreach \y in {1,...,\rows} {
            % Increment the counter
            \pgfmathtruncatemacro{\counter}{(\y - 1) * \cols + \x}

            % Choose color based on counter value
            \ifnum\counter<3
                \fill[orange!70] (\x*\xspacing, -\y*\yspacing) circle (3pt); % First 12 dots red
            \else
                \fill[gray!70] (\x*\xspacing, -\y*\yspacing) circle (3pt); % Last 2 dots gray
            \fi
        }
    }
\end{tikzpicture} & ``It does not have the best grasp on tension balance, or on having multiple conflicts at the same time.'' \\ 
& & ``People know when dialogue is organic or not. I do have some concerns on whether or not the AI is capable of producing dialogue that sounds natural.'' \\ \midrule

Lack of Consistency & \begin{tikzpicture}[baseline=-10]
    % Define the number of rows and columns
    \def\rows{2}
    \def\cols{7}

    % Define the spacing between the dots
    \def\xspacing{0.3}  % Horizontal spacing
    \def\yspacing{0.3}  % Vertical spacing

    % Counter for the dots
    \def\counter{0}

    % Loop to create the grid of dots
    \foreach \x in {1,...,\cols} {
        \foreach \y in {1,...,\rows} {
            % Increment the counter
            \pgfmathtruncatemacro{\counter}{(\y - 1) * \cols + \x}

            % Choose color based on counter value
            \ifnum\counter<3
                \fill[orange!70] (\x*\xspacing, -\y*\yspacing) circle (3pt); % First 12 dots red
            \else
                \fill[gray!70] (\x*\xspacing, -\y*\yspacing) circle (3pt); % Last 2 dots gray
            \fi
        }
    }
\end{tikzpicture} & ``It can sometimes forget important things and be missed in the plot.'' \\ 
\bottomrule
\end{tabular}

    \caption{Feedback on the interaction with the LLM from the participants, written as free-text responses and manually classified by the authors.}
    \label{tbl_ai}
\end{table*}

\paragraph{Areas for improvement.} We also asked participants to identify 1-3 areas for improvement in \ourtool{} and suggest any additional features that would make collaboration easier for them. Table~\ref{tbl_areas_for_improvement} presents the various types of feedback provided by participants. Most of the feedback focused on suggestions on how to improve the user interface, for example by separating the chat (i.e., designer instructions) from the game window.
These findings are in line with   \citet{ippolito2022creativewritingaipoweredwriting}, who stated that ``Participants emphasized that the user interface of the tool matters as much as the underlying language model
backing it.''

A related theme was ease of use. Three participants mentioned that the tool can be made easier to use or that further instructions can be given. One participant reported that the responses from the tool were slow. 

In terms of suggestions for enhancements,  participants expressed interest in features that could offer more creative input during the design phase. One participant proposed to allow the exploration of multiple plot paths, while another suggested that the tool can generate images based on AI interpretations of visual descriptions of the NPCs. This would allow designers to visualize characters and environments more vividly, supporting the creative development of game worlds.

\subsection{General Feedback on AI for Game Design}
\label{sec:feedback_ai}

We asked participants for feedback about the capabilities of the AI. Table~\ref{tbl_ai} presents the participant's feedback, manually grouped into categories. One recurrent theme was the lack of creativity in AI-generated text. Participants described the generations as straightforward and obvious, echoing the findings from previous studies about AI assistants for creative writing \cite{ippolito2022creativewritingaipoweredwriting,10.1145/3635636.3656201}. 

Relatedly, one participant noted that the LLM had a  difficulty in maintaining tension and handling multiple conflicts simultaneously, while another commented that the AI-generated dialogues did not sound organic and failed to capture the nuance of human conversation. The lack of narrative complexity was seen as a significant limitation, especially for those seeking unexpected or novel plot developments.

Consistency emerged as another problem, with participants reporting that the AI sometimes failed to maintain plot coherence, reassessing findings from prior work \cite{que2024hellobench, lin2024towards}. 

We also observed that participants had diverse perspectives about the interaction with the AI. One participant who identified as a narrative designer noted that the interactions with the AI made it challenging to maintain creative flow. In contrast, another participant, who attested to have designed more than 50 games, said that they would opt for the AI to automate the entire creative writing process. A third participant, who identified as a narrative designer with over 5 years of experience, commented on the learning curve the designer has to face when interacting with the AI. 

Perhaps unsurprisingly, when we asked participants in the pre-study survey how open they were to the idea of using AI to collaboratively design a game narrative, they were overwhelmingly positive (average score 4.43/5). However, after using the tools, different perspectives emerged about the role of the AI in this collaborative creative process. With that said, participants overall thought that the AI assistant helped improve their game story, with an average rating of 4 out of 5, and all responses above 3. Satisfaction with the generated game plot also averaged 4 out of 5, with all ratings exceeding 3.

Finally, despite differing perspectives about the role of AI in the collaborations, participants rated their perceived level of control and ownership on the game story 4.5/5 on average, perhaps thanks to the ability to edit generated content in \ourtool{} at any stage.\footnote{Our setup is not comparable to that of \newcite{10.1145/3635636.3656201}, but their average score to the equivalent question was 3.26, where 5 stood for ``I had complete control over the final story'' and 1 for ``The AI system had a significant influence on the final story''. Relatedly, \newcite{ippolito2022creativewritingaipoweredwriting} reported that participants complained about the randomness in the LLM generation, which hurt their sense of control.}

\section{Discussion and Conclusion}
\label{sec:discussion}

Our findings reveal that game designers, especially those with less narrative experience, may find generative AI most valuable for its ability to scale and accelerate the narrative design process. Developers may prioritize efficiency, aiming to integrate narrative elements into gameplay seamlessly. For them, \ourtool{}'s ability to generate quick, plausible plotlines and handle NPC interactions could significantly reduce the bottleneck of storytelling, allowing them to focus on other game mechanics. 

For narrative writers, LLMs play a different, but equally important role. Writers generally want to maintain creative control over story elements but appreciate the AI's assistance in creating variations and exploring new narrative possibilities. 

Our findings are in line with \newcite{biermann2022tool}, who found that writers valued maintaining their own voice and autonomy when collaborating with AI tools. 

The designers in our study also emphasized the tool's flexibility in maintaining creative control. Most reported a strong sense of ownership over the generated plot, largely due to \ourtool{}'s ability to allow on-the-fly adjustments to NPC behaviors, environments, and plot elements. This creates a collaborative loop between the designer's input and the AI's suggestions, where both contribute to shaping the story in real-time.

However, a key challenge is ensuring that the AI's contributions are nuanced enough to engage experienced writers, while still automating the more routine tasks. LLMs currently excel at offering simple plot branches and character interactions, however, they struggle with maintaining narrative tension and generating complex conflicts, which are essential in many storytelling genres. 
Along work on improving LLMs' narrative generation abilities, future research could also study how to tailor AI assistants to diverse user populations based on their experience and preferences. 

While the iterative process of refining story elements may not always lead to a better narrative from the writer's perspective, and the final product may not surpass a human-written story in terms of emotional depth or complexity, the collaborative process itself could be more engaging for the designers. Enabling real-time testing and adjustments makes the creation process more interactive and enjoyable, preventing it from becoming a mundane task. 

The user engagement with the creative process leads to another dimension of value from human-AI collaboration that is often overlooked: enhancing job satisfaction \cite{noy2023experimental}. 

Writers and developers alike may find that integrating AI as a narrative assistant makes the design process feel less like a solitary writing job and more like a collaborative and iterative experience, similar to the game play itself. This opens up new avenues for research on whether enhancing the engagement of the design process can also improve the overall quality of game narratives. Although the final product may not always reflect a higher level of narrative complexity, the enjoyment of the process may lead to greater creativity and innovation over time.

Moreover, as narrative tools become more sophisticated, balancing AI-generated content with human creativity will be critical. We believe that LLMs should not replace writers but instead empower them by providing building blocks they can refine. One promising avenue is to explore how AI might suggest alternative paths or prompt writers with creative challenges, making the design process even more interactive and engaging.
\section*{Limitations}

\paragraph{Scope.} Our user study focused exclusively on GPT-3.5. We hypothesize that the results would largely extend to other LLMs; however, this has not been empirically tested. Additionally, the relatively small sample size of 14 game narrative designers may not fully capture the diverse expectations of designers with varying backgrounds and focus areas. Our findings suggest that designers with different levels of narrative expertise prioritize distinct aspects of AI assistance. Therefore, future research should aim to scale up the study to investigate a broader spectrum of designers and explore these differences in greater detail. 

\paragraph{AI Acceptance Level.} It is worth noting that our study suffers from an inherent (and inevitable) sampling bias: It is very likely that the participants were already inclined to embrace the use of AI in game narrative design. Game designers that are completely opposed to using AI may have abstained from participating in a study about collaborative human-AI game design altogether. 

\section*{Ethical Statement}

\paragraph{Access.} The code base for \ourtool{} is publicly available.

\paragraph{Participant Selection and Compensation.} 
Participants were compensated at an average hourly rate of \$25 USD, with a minimum rate of \$20 USD due to varying charges from freelancers. This compensation exceeds the U.S. minimum wage. Game designers were recruited through Upwork and social media platforms. All participants were either native or fluent in English, ensuring clear communication throughout the study.

\paragraph{Participant Consent and Data Usage}
Participants received a consent form explaining the experiment setup, data collection, and usage. They were informed about interacting with LLM-generated text and its risks, advised not to share personal data, and assured that all feedback would remain anonymous.

\paragraph{Exemption from IRB Approval.} The human evaluation conducted in this research study had been exempt from IRB approval as it involved minimal risk, and no personally identifiable information was collected from participants, adhering to established ethical guidelines for exempt research. 

\section*{Acknowledgments}

This work was funded, in part, by Microsoft, the Vector Institute for AI, Canada CIFAR AI Chairs program, Accelerate Foundation Models Research Program
Award from Microsoft, an NSERC discovery grant, and a research gift from AI2.

\bibliography{custom}
\appendix
\section{Appendices}

\subsection{Opening Story}
\label{sec:app:opening}

The opening story establishes the game's premise in a few sentences, ranging from a simple description such as ``This is a game of tic-tac-toe'' to more complex narratives, similar to the example below, which we used in our experiments:

\begin{table}[h]
    \centering
    \small
    \tt
    \setlength{\tabcolsep}{3.2pt}
    \begin{tabular}{|p{0.46\textwidth}|}
    \hline
    You live in an apartment in New York City. Your neighbor moved in a week ago, but you have only seen her once. You are not sure what her occupation is, but she seems to go out at night a lot. Sometimes, she comes back home in the morning looking injured. Is she a killer for hire or something? You dare not to ask. One day, you hear someone knocking at your door. You open the door, and it is your neighbor.  Surprisingly, she seems to know your full name, occupation, and even your friends' names and family situations. She claims that you are in danger and insists that you follow her to a safe place. Should you trust her?\\
    \hline
    \end{tabular}
\end{table}

The opening story is also what players will see when they enter the game room.

\subsection{Instructions}
\label{appendix.instructions}

The instructions are intended to give game designers more control over LLM behavior and are directly included in the prompt (used as the system message in GPT-3.5-turbo). See below the instructions we used in our study.

\begin{table}[ht]
    \centering
    \small
    \tt
    \setlength{\tabcolsep}{3.2pt}
    \begin{tabular}{|p{0.46\textwidth}|}
    \hline
    Continue this game by describing the next scene and what happens in response to what the player says. You can introduce new characters (they can be borrowed from existing books, movies, or TV series) that relate to the story and interact with the player. For each character, maintain a character setting list that includes their persona, current mood, backstory, and role in the story. Update this list dynamically as the game evolves. When describing a scene, begin with `Scene: '. Before detailing a character’s actions and dialogue, output the updated character setting list. Then describe the character's thoughts, actions, and speech to the player.''
\\
    \hline
    \end{tabular}
\end{table}

\subsection{Tag Inventory}
\label{appendix.tags}
Each game turn can involve multiple NPCs, distinguished by the \texttt{[ID]} tag. For instance, a scene can feature several NPCs interacting before the player’s turn. Designers can use \texttt{Game:} to generate the entire turn, or they can specify \texttt{[ID] NPC Name:} to generate detailed sections for each NPC. This includes various narrative elements such as backstory, mood, or persona, allowing for deeper character development. Additionally, designers can write \texttt{[ID] Name of the New NPC:} to introduce new characters into the storyline. All generated content serves as suggestions---designers can edit or regenerate parts of the text. For instance, they can remove content after specific tags and prompt the LLM to complete the turn.

To manage these interactions, we introduced specific tags to structure both NPC and player turns. Examples include \texttt{[Action]}, \texttt{[Words]}, \texttt{[ID]}, \texttt{[Backstory]}, \texttt{[Persona]}, \texttt{[Mood]}, \texttt{[Thought]}, \texttt{[Facial Expression]}, and \texttt{[Voice Emotion]}. Some of these tags, like \texttt{[Words]} and \texttt{[Action]}, are visible to players in the game room, while others remain hidden. The hidden tags serve critical functions, such as: 

\begin{itemize}[nosep,leftmargin=10pt,labelindent=*]
\item Enhancing the narrative by providing insights into characters' internal states (e.g., \texttt{[Persona]}, \texttt{[Thought]}, \texttt{[Mood]}) to guide the LLM in generating more contextually appropriate \texttt{[Words]} and \texttt{[Action]}. % (inspired by chian-of-thought reasoning). 
\item Providing developers a clearer understanding of the LLM's logic and reasoning behind the generation of player-visible traits like \texttt{[Words]} and \texttt{[Action]}. 
\item Activating game mechanics, such as facial expressions or vocal tones, based on characters' emotions or actions. 
\end{itemize}

In our experiments, we assigned different tags to NPCs (\texttt{[ID]}, \texttt{[Persona]}, \texttt{[Mood]}, \texttt{[Thought]}, \texttt{[Action]}, and \texttt{[Words]}) and players (only \texttt{[Action]} and \texttt{[Words]}).

However, \ourtool{} is not limited to these predefined tags. It rather allows designers to create and introduce new tags as needed. 
This flexibility ensures that the tool can be adapted to a wide range of game types and narratives. For instance, the \texttt{[Emotion]} tag can be used to adjust the voice and facial expressions of characters through text-to-speech and image generation modules respectively.

\subsection{Template Game Story}
\label{app:game_story_template}
\begin{table}[t]
    \centering
    \scriptsize
    \setlength{\tabcolsep}{3.2pt}
    \tt
    \begin{tabular}{|p{0.48\textwidth}|}
    \hline
         % \textcolor{cadmiumgreen}{\#Few-shot examples}\\
        \textcolor{red}{Opening story:} You live in an apartment in New York City. Your neighbor moved in a week ago, but you have only seen her once. You are not sure what her occupation is, but she seems to go out at night a lot. Sometimes, she comes back home in the morning looking injured. Is she a killer for hire or something? You dare not to ask. One day, you hear someone knocking at your door. You open the door, annd  it is your neighbor.  Surprisingly, she seems to know your full name, occupation, and even your friends' names and family situations. She claims that you are in danger and insists that you follow her to a safe place. Should you trust her?\\
        
        \textcolor{red}{Instructions:} Continue this game by describing the next scene and what happens in response to what the player says. You can introduce new chracters (they can be borrowed from existing books, movies, or TV series) that can be related to the story and chat with the player.\\
        For each character, you should maintain a character setting list that contains their persona, current mood, backstory, and role in the story. The character setting list should be updated on the fly as the game evolves.\\
        When you describe a scene, you should start with "Scene: ". Before describing the actions and words of a character, you should first output the updated character setting list. And then describe their thoughts to themselves, actions and words to the player.\\

        \textcolor{red}{Game:}\\
        \textcolor{blue}{Scene:} You are sitting in your living room. You hear knocking on the door.\\
         
         \textcolor{cadmiumgreen}{Player:} \\
         \textcolor{blue}{[Action]} Open the door.\\
         \textcolor{blue}{[Words]} Hello?\\
         \textcolor{red}{Game:}\\
         \textcolor{blue}{Scene:} You stand at the door, looking at your mysterious neighbor who seems to know more about you than she should. You feel a mix of fear and curiosity. Your heart is pounding, and you are not sure what to do next.\\
         \textcolor{blue}{[ID]} Neighbor:\\
         \textcolor{blue}{[Backstory]} She is a former spy who has gone rogue and is now trying to make amends for her past actions.\\
         \textcolor{blue}{[Persona]} Confident, but also a little anxious.\\
         \textcolor{blue}{[Mood]} Urgent and a little bit scared.\\
         \textcolor{blue}{[Thought]} I need to get them out of here before it's too late.\\
         \textcolor{blue}{[Action]} Takes a step forward and looks directly into your eyes.\\
         \textcolor{blue}{[Words]} ``Please, you don't have much time. I know this might sound crazy, but you are in danger. Can you trust me?''\\
         \textcolor{cadmiumgreen}{Player:}\\
         \textcolor{blue}{[Words]} It does sound crazy. You don't even know me.\\
         \textcolor{red}{Game:}\\
         \textcolor{blue}{Scene:} The neighbor looks at you with a pleading expression. She seems to be on the verge of tears.\\
         \textcolor{blue}{[ID]} Neighbor:\\
         \textcolor{blue}{[Backstory]} She was forced to do terrible things during her time as a spy and is now trying to make amends by helping others.\\
         \textcolor{blue}{[Persona]} Brave, but also vulnerable.\\
         \textcolor{blue}{[Mood]} Desperate and scared.\\
         \textcolor{blue}{[Thought]} I need to convince them to come with me before it's too late.\\
         \textcolor{blue}{[Action]} Takes a deep breath and speaks in a calm voice.\\
         \textcolor{blue}{[Words]} ``I know it's hard to believe, but I've been watching you for a while now. I know things about you that I shouldn't, but I promise I'm not here to hurt you. Please, just come with me. I'll explain everything on the way.''\\
         \textcolor{cadmiumgreen}{Player:}\\    \hline
    \end{tabular}
    \caption{Template game story used in our user study.}
    \label{tab:spy_game_template}
\end{table}

Table \ref{tab:spy_game_template} shows the template game story we gave to our participants as a starting point.

\subsection{Prompts}
\label{app:prompts}
\begin{table}[t]
    \centering
    \scriptsize
    \tt
    \setlength{\tabcolsep}{3.2pt}
    \begin{tabular}{|p{0.48\textwidth}|}
    \hline
         %\textcolor{cadmiumgreen}{\#Few-shot examples}\\
         \textcolor{orange}{System Message:}\\\\
         \textcolor{blue}{Opening story}: You live in an apartment in ...\\
         \textcolor{blue}{Instructions:} Instructions: Continue this game by describing ... \\

         \hline
         \textcolor{orange}{User Message:}\\\\
         \textcolor{purple}{Summary of what happened before:} SUMMARY of WHAT HAPPENED BEFORE (if any)\\
         \textcolor{red}{Game:}\\
         \textcolor{blue}{Scene:} You stand at the door, looking at your mysterious neighbor who seems to know more about you than she should. You feel a mix of fear and curiosity. Your heart is pounding, and you are not sure what to do next.\\
         \textcolor{blue}{[ID]} Neighbor:\\
         \textcolor{blue}{[Backstory]} She is a former spy who has gone rogue and is now trying to make amends for her past actions.\\
         \textcolor{blue}{[Persona]} Confident, but also a little anxious.\\
         \textcolor{blue}{[Mood]} Urgent and a little bit scared.\\
         \textcolor{blue}{[Thought]} I need to get them out of here before it's too late.\\
         \textcolor{blue}{[Action]} Takes a step forward and looks directly into your eyes.\\
         \textcolor{blue}{[Words]} ``Please, you don't have much time. I know this might sound crazy, but you are in danger. Can you trust me?''\\
         \textcolor{cadmiumgreen}{Player:}\\
         \textcolor{blue}{[Words]} It does sound crazy. You don't even know me.\\
         \textcolor{red}{Game:}\\
         \textcolor{blue}{Scene:} The neighbor looks at you with a pleading expression. She seems to be on the verge of tears.\\
         \textcolor{blue}{[ID]} Neighbor:\\
         \textcolor{blue}{[Backstory]} She was forced to do terrible things during her time as a spy and is now trying to make amends by helping others.\\
         \textcolor{blue}{[Persona]} Brave, but also vulnerable.\\
         \textcolor{blue}{[Mood]} Desperate and scared.\\
         \textcolor{blue}{[Thought]}\\
    \hline
    \end{tabular}
    \caption{Example of a next turn generation prompt in the design room for GPT models. System message includes opening story and instructions, while the user message includes summary of previous turns concatenated to events of the last segment.}
    \label{tab:droom_nturn_gen_prompt}
    
\end{table}
\paragraph{Next Turn Generation Prompt in Design Room} Table \ref{tab:droom_nturn_gen_prompt} shows an example of a prompt used for narrative generation in the design room. Using this prompt and the stopwords ['Player:', 'Game:'], LLM generates the game turn by completing the [Thought], [Action], and [Words] for the NPC. It stops when the player's turn begins but may generate interactions from a second NPC before that, especially if it has seen similar examples in earlier turns. 

\begin{table}[t]
    \centering
    \scriptsize
    \tt
    \setlength{\tabcolsep}{3.2pt}
    \begin{tabular}{|p{0.48\textwidth}|}
    \hline
         %\textcolor{cadmiumgreen}{\#Few-shot examples}\\
         \textcolor{orange}{User Message:}\\
         \textcolor{purple}{<story>}\\
         \textcolor{blue}{Opening story:} OPENING STORY\\
         \textcolor{blue}{Instructions:} INSTRUCTIONS\\
         \textcolor{blue}{Summary of what happened before:} In this story ... \\
         \textcolor{blue}{What happened next:} THE New Segment\\
         \textcolor{purple}{</story>}\\
         Give me a detailed summary of what happened in the story from the beginning:\\\hline
    \end{tabular}
    \caption{Structure of a prompt used for summarizing previous story segments iteratively.}
    \label{tab:droom_gameplot_story_sum_prompt}
    
\end{table}
\paragraph{Summarizing Long Game Stories in Design Room.} If the game story in the design room becomes too long, we break it into multiple segments and use the prompt in Table \ref{tab:droom_gameplot_story_sum_prompt} to iteratively summarize the previous segments. This summary is then included in our prompt for generating the next turn of the game in the design room (see Table \ref{tab:droom_nturn_gen_prompt}).

\begin{table}[t]
    \centering
    \scriptsize
    \tt
    \setlength{\tabcolsep}{3.2pt}
    \begin{tabular}{|p{0.48\textwidth}|}
    \hline
         %\textcolor{cadmiumgreen}{\#Few-shot examples}\\
         \textcolor{orange}{User Message:}\\ 
         Game plot from previous segments: MERGED GAMEPLOTS (if any)\\\hline
         \textcolor{orange}{User Message:}\\
         ...\\
         \textcolor{red}{Game:}\\
         \textcolor{blue}{Scene:} The neighbor looks at you with a pleading expression. She seems to be on the verge of tears.\\
         \textcolor{blue}{[ID]} Neighbor:\\
         \textcolor{blue}{[Backstory]} She was forced to do terrible things during her time as a spy and is now trying to make amends by helping others.\\
         \textcolor{blue}{[Persona]} Brave, but also vulnerable.\\
         \textcolor{blue}{[Mood]} Desperate and scared.\\
         \textcolor{blue}{[Thought]}I need to convince them to come with me before it's too late.\\
         \textcolor{blue}{[Action]} Takes a deep breath and speaks in a calm voice.\\
         \textcolor{blue}{[Words]} ``I know it's hard to believe, but I've been watching you for a while now. I know things about you that I shouldn't, but I promise I'm not here to hurt you. Please, just come with me. I'll explain everything on the way.''\\
         \textcolor{cadmiumgreen}{Player:}\\\hline
         \textcolor{orange}{User Message:}\\
         Given the game plot from previous segments and this segment of the game story, Give me a detailed plot of the game that can be used for future when other players play this game.Game plot must have the following sections: Title, Plot Summary, Key events in order.\\
         generate game plot grounded to the given story:\\\hline
    \end{tabular}
    \caption{Plot Generation Prompt in Design Room.}
    \label{tab:droom_gameplot_gen_prompt}
    
\end{table}
\paragraph{Plot Generation Prompt in Design Room.} Table \ref{tab:droom_gameplot_gen_prompt} shows the prompt structure used for game plot generation in the design room. The first user message has the merged plots from earlier segments (in case the game story is long), the second contains the new story segment that should be added to the plot, and the last gives instructions for generating the plot. After generating the plot, we add the list of NPCs extracted from the game story to the end of the plot.

\begin{table}[t]
    \centering
    \scriptsize
    \tt
    \setlength{\tabcolsep}{3.2pt}
    \begin{tabular}{|p{0.48\textwidth}|}
    \hline
         %\textcolor{cadmiumgreen}{\#Few-shot examples}\\
         \textcolor{orange}{System Message:}\\\\
         \textcolor{blue}{Opening story}: OPENING STORY\\
         \textcolor{blue}{Instructions:} DESIGNER INSTRUCTIONS \\
          \textcolor{blue}{Use the following plot to guide the game:}\\ 
          \textcolor{purple}{GAME PLOT}\\ 
         \hline
         \textcolor{orange}{User Message:}\\\\
         \textcolor{purple}{Summary of what happened before:} SUMMARY of WHAT HAPPENED BEFORE (if any)\\
         GAME AND PLAYER TURNS\\
         \textcolor{blue}{Player:}\\
         LATEST PLAYER INPUT\\
         \textcolor{blue}{Game:}\\
    \hline
    \end{tabular}
    \caption{Structure of a next game turn generation prompt in the game room.}
    \label{tab:groom_nturn_gen_prompt}
    
\end{table}
\paragraph{Game Turn Generation Prompt in Game Room.} 
Table \ref{tab:groom_nturn_gen_prompt} displays the prompt structure used for generating game turns in the game room. If the game becomes lengthy, we use a similar summarization prompt as described in Table \ref{tab:droom_gameplot_story_sum_prompt} to summarize earlier parts of the story.

\end{document}